\documentclass{article}
\usepackage[preprint]{neurips_2026}
\usepackage[T1]{fontenc}
\usepackage{newtxtext,newtxmath}
\usepackage{microtype}
\usepackage{graphicx}
\usepackage{booktabs}
\usepackage[hidelinks]{hyperref}

\title{Computer-Use Agents Are Embodied Generalists}
\author{Pranjal Aggarwal \qquad Sean Welleck\\Carnegie Mellon University}
\date{}

\begin{document}
\maketitle

\begin{abstract}
In many areas of AI, general-purpose models have increasingly replaced handcrafted methods and specialized architectures. We argue that embodied AI is poised for a similar shift. Our key claim is that embodied tasks can be formulated as interaction with digital software. Specifically, we build software that renders an agent's view of a physical environment on a screen and provides controls for acting in it. A computer-use agent (CUA) can then perform embodied tasks through the same screenshots and keyboard and mouse actions that it uses to operate ordinary software. Using a single CUA design and changing only the software, our entries win first-, second-, third-, and sixth-place prizes across four ECCV workshop competitions, without any competition-specific agent tuning. These results suggest that the skills CUAs learn on software, such as visual grounding, precise clicking, and exploration, already carry over to embodied tasks, giving CUAs a far broader scope in embodied AI than previously realized. We therefore advocate that computer-use agents can be embodied generalists. Further, we advocate that, by following the training strategy that made computer-use agents general, embodied AI can converge on a single general model as well, ushering in a new era of general embodied AI.\footnote{Website: \url{https://pranjal2041.github.io/cua-embodied}}
\end{abstract}

\suppressfloats[t]
\begin{figure}[t]
    \centering
    \includegraphics[width=\linewidth]{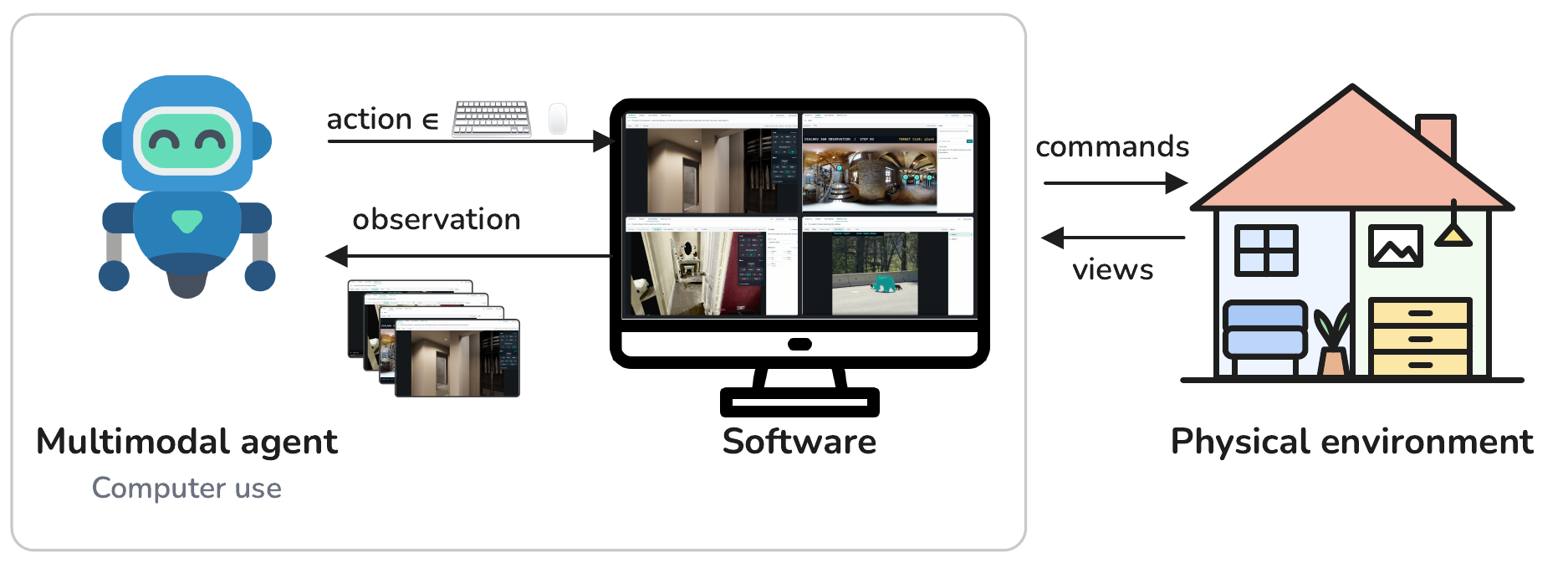}
    \caption{Computer-use agents can operate any software through keyboard and mouse actions while observing the screen. We argue that physical environments are no different. Once exposed through software, they become just another interface for the same agent.}
    \label{fig:teaser}
\end{figure}

\section{Introduction}

In many areas of AI, general-purpose models have increasingly replaced handcrafted methods and specialized architectures~\citep{sutton2019bitter,elkishky2025competitive,luong2025imogold}. We argue that embodied AI, which studies agents that perceive and act in physical environments, is poised for a similar shift~\citep{oneill2024openx,gemini2025robotics}. In particular, we explore connecting embodied AI with computer-use agents: general-purpose digital agents that operate diverse software applications through screenshots, mouse clicks, and keyboard inputs. Although operating a computer and acting in a physical environment may seem quite different, we observe that both involve similar challenges, including visual grounding, exploration, and adapting actions based on feedback from the environment~\citep{yang2025magma,zhang2025appagent,huang2022innermonologue}.

Our key insight is that embodied tasks can be formulated as interaction with digital software. Specifically, we construct software that renders observations on a screen and provides controls for acting in the physical environment (Figure~\ref{fig:teaser}). A computer-use agent can then observe the environment, take actions, and inspect their consequences using the same interface that it uses to operate ordinary desktop software. Task-specific functionality resides in the software rather than the agent design, allowing us to use existing computer-use agents across different embodied tasks without developing specialized models.

To investigate our approach, we develop software applications for four embodied AI tasks: vision-language navigation, dialogue-driven navigation, language-guided 3D segmentation, and referring video segmentation. We use a single, minimal computer-use agent harness across all four applications, without competition-specific agent tuning, and enter 4 ECCV 2026 workshop competitions. Our entries place first in OpenSUN3D, second in MeViSv2-Text, third in DialNav, and sixth in the VLNVerse competitions. These results demonstrate that computer-use agents can achieve strong performance across diverse embodied tasks by interacting with task-specific software through a shared agent design.

Beyond these results, we argue that embodied AI can follow the same path that computer use has already taken. Only two years ago, computer-use agents faced the challenges that embodied AI faces today. They struggled to click precisely and relied on overlays or separate grounding models to locate targets~\citep{yang2023som,zheng2024seeact,gou2025uground}, much as most VLAs today add a dedicated action head for continuous control~\citep{black2025pi05,nvidia2025gr00t}. They also lagged far behind agents with task-specific APIs~\citep{xie2024osworld,aggarwal2026pwp,song2025beyondbrowsing}. Today, the same model that writes code and solves mathematics problems can also operate computers by writing click coordinates as ordinary text~\citep{anthropic2024computeruse,google2026gemini35cu}, and the strongest agents perform comparably whether they act through the screen or through APIs~\citep{neocognition2026apprenticebench}. This progress came from training for precise visual grounding, building environments for reinforcement learning, and making interaction fast enough for real-time use~\citep{gou2025uground,aggarwal2026gymanything,wang2026cuagym,aggarwal2026cuaspeedrun}. Each of these has a direct counterpart in embodied AI. Simulators can generate grounding data at scale~\citep{yuan2024robopoint}, embodied environments can be used for reinforcement learning~\citep{li2026simplevlarl}, and faster inference can enable real-time control~\citep{black2025rtc}. Since our approach exposes embodied environments through software, advances in computer-use agents can also transfer to embodied AI directly, and \textit{embodied tasks can become part of the training of future general-purpose agents}. We therefore believe that embodied AI will converge on a single general model, just as computer use did~\citep{openai2026astra,anthropic2026opus55}.

\section{Methodology}
\label{sec:method}

Our goal is to provide sparks of evidence for our hypothesis that the capabilities computer-use agents learn on digital software also generalize to embodied environments. We entered four different ECCV workshop competitions, and evaluated a single computer-use agent on their official test sets. We first describe the computer-use agent shared across all four, and then the software we built for each competition.

\subsection{A shared computer-use agent}

We keep the computer-use agent's design fixed and construct software for each competition. The agent receives the task description, the software's instructions, and any reference images or videos. Let $q$ denote these task inputs, $\mathcal{S}_k$ the software for competition $k$, and $s_t$ its state at step $t$. The software presents a visual observation $o_t$, consisting of a screenshot and any video returned by an operation. Given the interaction history $h_t=(o_0,a_0,\ldots,o_t)$, the agent $\pi$ chooses a keyboard or mouse action $a_t$:
\begin{equation}
    a_t \sim \pi(\cdot\mid q,h_t),
    \qquad
    (s_{t+1},o_{t+1}) = \mathcal{S}_k(s_t,a_t).
\end{equation}
The software executes the action and returns the updated observation. The agent repeats this loop until it presses submit through the interface, at which point the software exports the trajectory or prediction.

\begin{figure}[t]
    \centering
    \begin{minipage}[t]{0.49\linewidth}
        \centering
        \includegraphics[width=\linewidth]{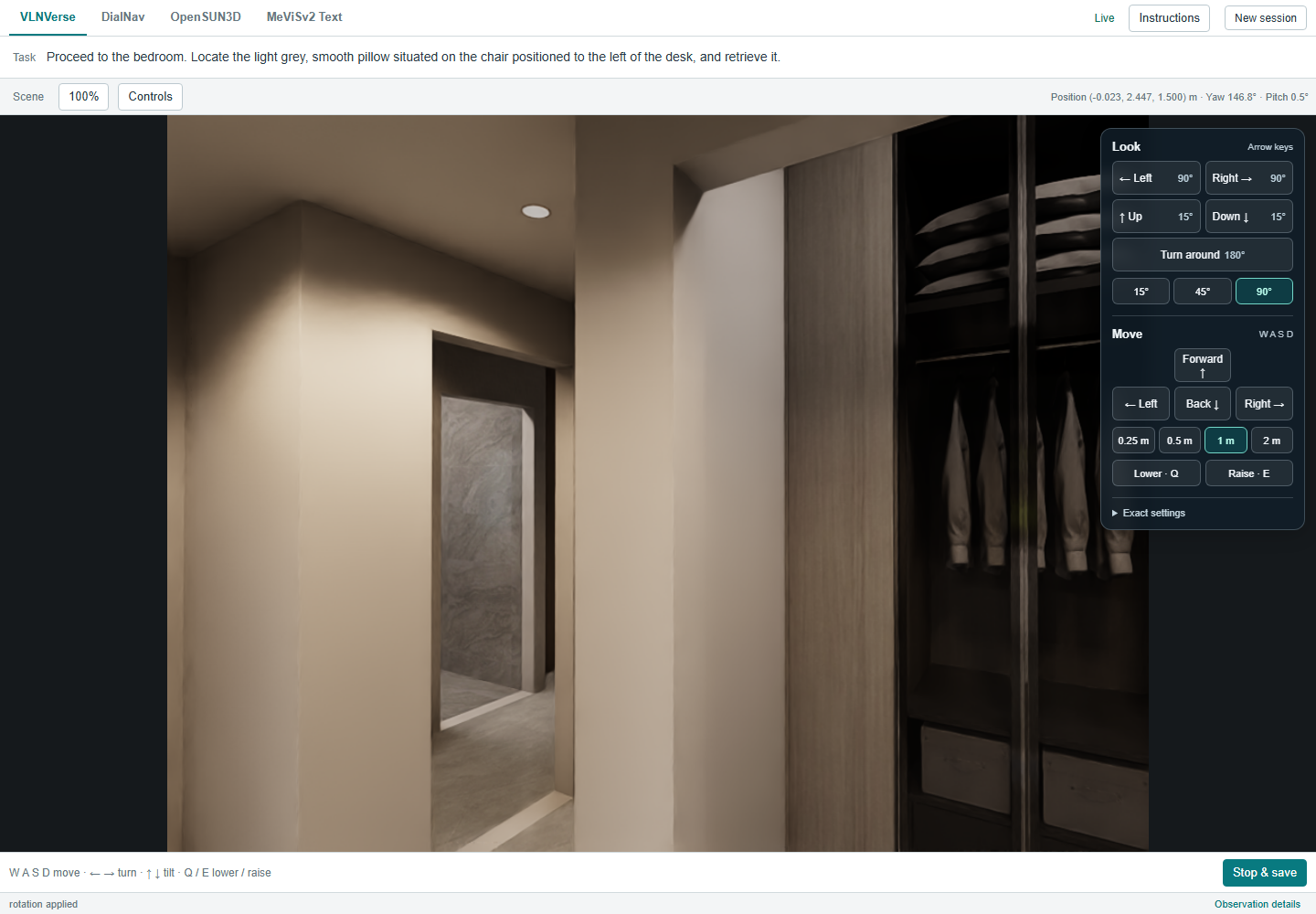}\\[3pt]
        {\small (a) VLNVerse: move and look around.}
    \end{minipage}\hfill
    \begin{minipage}[t]{0.49\linewidth}
        \centering
        \includegraphics[width=\linewidth]{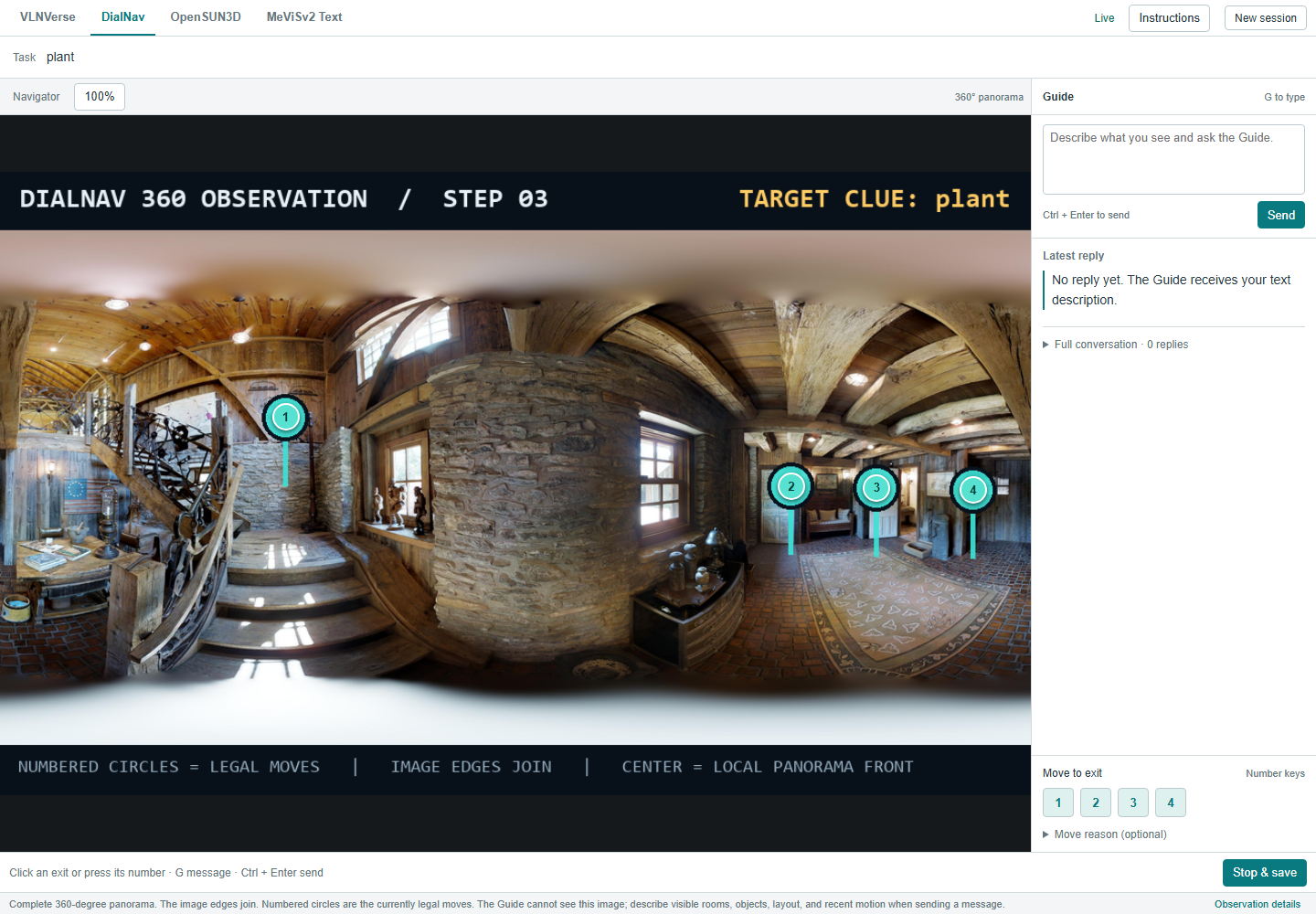}\\[3pt]
        {\small (b) DialNav: ask the Guide and choose exits.}
    \end{minipage}

    \vspace{9pt}
    \begin{minipage}[t]{0.49\linewidth}
        \centering
        \includegraphics[width=\linewidth]{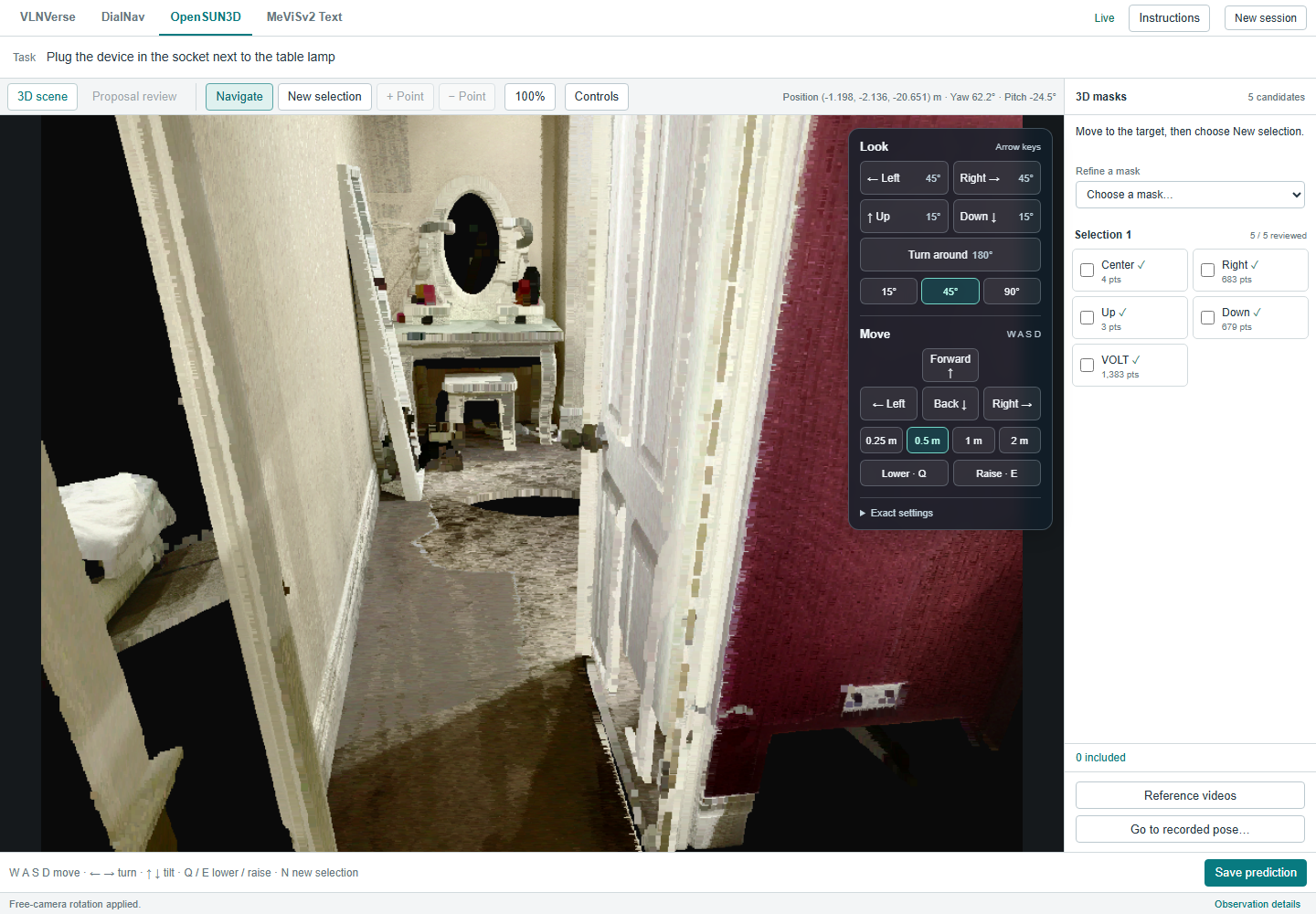}\\[3pt]
        {\small (c) OpenSUN3D: explore and select 3D regions.}
    \end{minipage}\hfill
    \begin{minipage}[t]{0.49\linewidth}
        \centering
        \includegraphics[width=\linewidth]{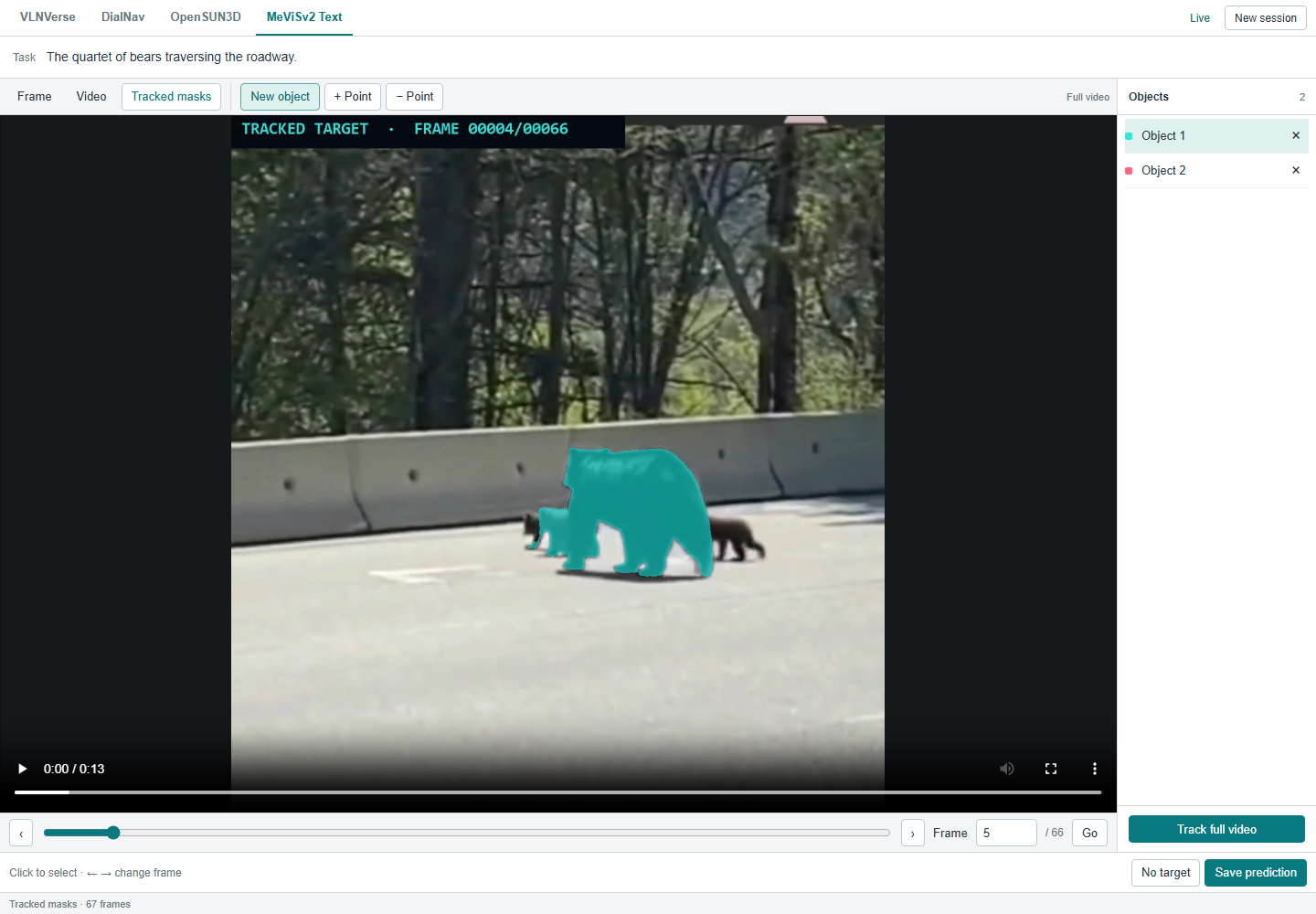}\\[3pt]
        {\small (d) MeViSv2: select objects and review tracks.}
    \end{minipage}
    \caption{The shared computer-use agent operates four software applications through keyboard and mouse actions. Each application provides task-specific controls and displays the outcomes.}
    \label{fig:software}
\end{figure}

We use Gemini 3.5 Flash and Gemini 3.6 Flash with the same agent implementation, system prompt, and keyboard and mouse action space across the four software applications. Each application provides controls for its internal operations, including movement, dialogue, segmentation, and mask propagation. The agent chooses when to use these operations, interprets their outputs, and decides what to do next. It can issue individual inputs or batches of up to 16 inputs; each operation finishes before the next begins, and a screenshot follows the batch. We retain the interaction text and remove the oldest ten screenshots when the context reaches twenty. Complete reference videos and generated overlay videos remain in context. Figure~\ref{fig:software} shows the four interfaces.

\subsection{VLNVerse: navigation through a 3D scene}

VLNVerse~\citep{lin2025vlnverse} requires navigating to a destination described in language. We construct a 3D scene viewer with a first-person image and controls for translation and rotation. The agent moves forward, backward, sideways, or vertically using keyboard shortcuts or on-screen buttons. It can turn, look up or down, and adjust the movement distance and rotation angle. Each operation updates the camera pose and renders a new image. The agent uses these observations to identify landmarks, explore rooms, and approach the target. The same viewer supports the coarse and fine tracks. Selecting the stop control saves the trajectory and final pose.

\subsection{DialNav: navigation with a conversational guide}

DialNav~\citep{han2025dialnav} requires reaching a destination through dialogue with a Guide. Our software displays a $360^\circ$ panorama, numbered exits to adjacent viewpoints, a target clue, and a message panel. The agent types descriptions and questions, reads the Guide's replies, and clicks exits to move. A Guide model within the software receives the destination, a scene atlas, and route previews in a separate context. It infers the agent's location from its descriptions and returns natural-language directions. The agent uses these directions and the panorama to choose each movement. The stop control saves the trajectory and dialogue.

\subsection{OpenSUN3D: language-guided 3D segmentation}

The OpenSUN3D challenge~\citep{opensun3d2026} uses SceneFun3D~\citep{delitzas2024scenefun3d} tasks that describe an action and require a 3D mask of the corresponding interaction region, such as a handle for opening a drawer. Our software renders the colored Faro point cloud and provides free-camera movement, rotation, and point selection. Reference videos supply scene context and shortcuts to recorded camera poses. The agent explores the 3D scene, approaches the target, and clicks the relevant component. The renderer maps this click to a visible Faro point. Within the software, PinPoint3D~\citep{zhang2025pinpoint3d} produces point-prompted masks, while a VOLT model fine-tuned on SceneFun3D supplies affordance-based proposals. The interface displays each candidate from four viewpoints. The agent can refine a selection with positive or negative points, move to another viewpoint, or make a new selection. It then chooses which masks to save, retaining their original Faro point indices.

\subsection{MeViSv2: referring video segmentation}

The MeViSv2~\citep{ding2025mevis} text track requires segmenting the objects described by a motion expression throughout a video. Our software provides a video player, a frame selector, and controls for adding, refining, and removing objects. The agent receives the complete video and expression, chooses a frame, and clicks each target. SAM3~\citep{carion2025sam3} segments these objects and propagates their masks through the video. The agent receives the complete tracked-mask overlay and can inspect individual frames. It can add positive or negative points, remove an object, or select another object before running propagation again. The save control exports the masks of all remaining objects. A no-target control exports an empty mask sequence.

\section{Results}
\label{sec:results}

\subsection{Competition results}

The same computer-use agent, with only its software changed, won prizes in all four competitions, spanning navigation, dialogue-driven navigation, 3D segmentation, and video segmentation. It placed first in OpenSUN3D, second in MeViSv2-Text, third in DialNav, and sixth in VLNVerse. We report the official test scores of our competition entries, together with the best other entry in each competition, in Table~\ref{tab:competition-results}. In OpenSUN3D, the agent finished ahead of every other entry by more than 4 points in AP50 and 10 points in AP25. There, it explored a 3D scan and clicked the exact part to interact with, such as a drawer handle, much as it clicks a small button on a screen. In VLNVerse, the agent placed sixth on success but followed the reference routes more closely than all but two of the 74 entries, with the third-highest NDTW on both tracks. In MeViSv2-Text, it placed second ahead of an entry with a higher $\mathcal{J}\&\mathcal{F}$, because it more often recognized when no object matched the expression (88.89 versus 83.33 no-target accuracy). Each of these results rests on abilities the agent already uses on digital software, namely locating targets, pointing at them precisely, and exploring its environment.

\begin{table}[!ht]
    \centering
    \caption{Official test scores of our competition entries and the best other entry for each metric, taken from the official challenge results~\citep{emr2026,ead2026,opensun3d2026,liu2026lsvos}. All scores are expressed on a 0--100 scale.}
    \label{tab:competition-results}
    \begin{tabular}{@{}lllrr@{}}
        \toprule
        Competition & Place & Metric & Ours & Best other \\
        \midrule
        VLNVerse & 6th & Avg SR@3 & 87.30 & 90.70 \\
                 &     & NDTW (coarse / fine) & 95.0 / 96.8 & 96.3 / 97.6 \\
        DialNav & 3rd & Endpoint success rate & 65.26 & -- \\
                &     & SPL & 52.87 & -- \\
        OpenSUN3D & 1st & AP50 & 25.88 & 21.45 \\
                  &     & AP25 & 36.52 & 26.50 \\
        MeViSv2-Text & 2nd & Final & 86.72 & 90.81 \\
                     &     & $\mathcal{J}\&\mathcal{F}$ & 74.96 & 79.22 \\
        \bottomrule
    \end{tabular}
\end{table}

VLNVerse's SR@3 measures the fraction of episodes ending within three metres of the target. Avg SR@3 averages it over the coarse and fine tracks, and NDTW measures how closely a path follows the reference route. DialNav reports the endpoint success rate and success weighted by path length (SPL). OpenSUN3D evaluates 3D instance masks using average precision at 50\% and 25\% intersection over union. MeViSv2's Final score averages $\mathcal{J}\&\mathcal{F}$ with the accuracy of predicting target presence and absence.

\subsection{GUI operations or text APIs?}

A natural question is whether we need to construct GUIs or could instead provide direct text APIs for navigation and other operations. We view computer use as operating a digital interface. GUIs and APIs are two choices for that interface. We expect sufficiently capable computer-use agents to achieve the same task performance with either interface, although efficiency may differ.

We evaluate this by comparing GUI and API interaction on a subset of tasks from each of the four competitions. Each pair uses the same model, task inputs, and software operations. We compare successive generations of Gemini Flash to examine how improvements in computer use affect the performance gap between interfaces.

Table~\ref{tab:gui-api} shows the results on the validation sets. The two interfaces performed within four points of each other, and the GUI scored higher in two of the four competitions.

\begin{table}[!ht]
    \centering
    \caption{GUI and API interaction on the validation set of each competition, with the same model, task inputs, and software operations. All scores are expressed on a 0--100 scale.}
    \label{tab:gui-api}
    \begin{tabular}{@{}llrrr@{}}
        \toprule
        Competition & Metric & API & GUI & GUI $-$ API \\
        \midrule
        VLNVerse & Avg SR@3 & 83.9 & 85.2 & $+1.3$ \\
        DialNav & Endpoint success rate & 67.4 & 63.4 & $-4.0$ \\
        OpenSUN3D & AP50 & 26.7 & 28.1 & $+1.4$ \\
        MeViSv2-Text & Final & 95.8 & 94.6 & $-1.2$ \\
        \bottomrule
    \end{tabular}
\end{table}

This expectation is consistent with prior work. The performance gap between GUI-only and API-assisted interaction has shown to be narrowing across claude generations~\citep{aggarwal2026pwp}, while more recently, the strongest evaluated agents achieve comparable or higher task success through GUIs than through APIs~\citep{neocognition2026apprenticebench}.

\section{Position: Computer-use agents are embodied generalists}

\begin{figure}[t]
    \centering
    \includegraphics[width=\linewidth]{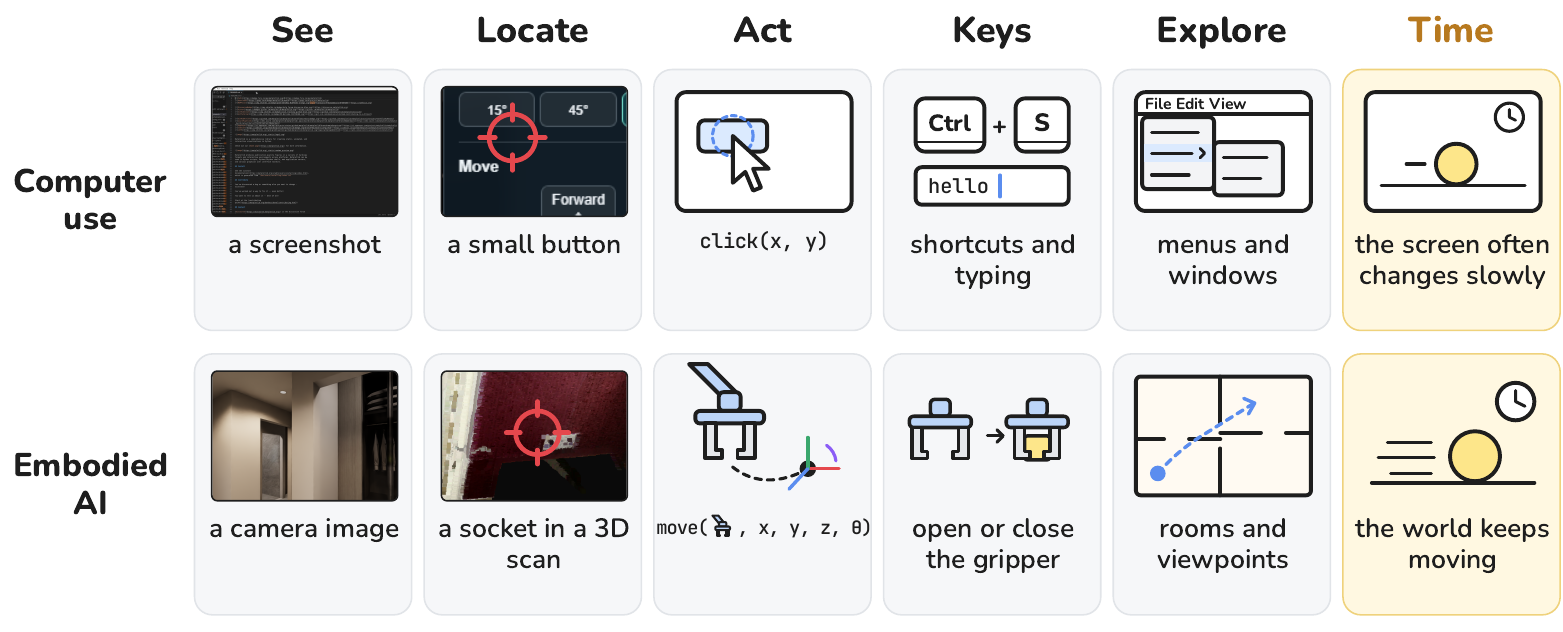}
    \caption{Computer use and embodied AI require much of the same from an agent. In both, it must see its surroundings, locate a target, act on it precisely, issue discrete commands, and explore. The main difference is time, as the screen often changes slowly while the physical world keeps moving.}
    \label{fig:analogy}
\end{figure}

\textbf{Claim 1: Representing physical interfaces in software lets us apply the methods that improved computer-use agents to embodied AI}

The challenges facing embodied AI/robotics today are similar to those faced by computer-use agents two years ago. Both require an agent to ground fine-grained actions in visual observations and adapt as its environment changes~\citep{cheng2024seeclick,yuan2024robopoint,lee2025molmoact,yang2025magma} (Figure~\ref{fig:analogy}). For computer use, this means locating a target on a screen and clicking at the correct coordinates. For robotics, it means locating targets in three dimensions and coordinating motor commands to achieve the intended positions and orientations. Representing physical interfaces in software or simulations lets us apply the same methods for training agents to perform these interactions. Our results in Section~\ref{sec:results} offer initial evidence: using the same agent and changing only the software, we performed competitively in four embodied competitions.

Fine-grained motor control and real-time physical interaction may seem out of reach for computer-use agents. Yet only two years ago, precise clicking seemed equally out of reach: computer-use agents struggled to operate digital interfaces~\citep{xie2024osworld} and lagged behind agents equipped with task-specific APIs~\citep{aggarwal2026pwp,song2025beyondbrowsing}. As discussed in the previous section, the strongest agents now achieve comparable task performance through either interface~\citep{neocognition2026apprenticebench}. We therefore argue for following the same development strategies in embodied AI, from learning precise, visually grounded actions to improving reliability and speed.

\textbf{Claim 2: Embodied AI will converge on a single general model, as computer use did}

Much of embodied AI today is built on vision-language-action models (VLAs), a special case of vision-language models further fine-tuned for robotic control~\citep{brohan2023rt2,kim2024openvla,black2025pi0,kawaharazuka2025vlareview}. Since robot actions are continuous, and discrete tokens seem too coarse to express them precisely, most VLAs also add a dedicated action head, typically trained with diffusion or flow matching~\citep{black2025pi0,nvidia2025gr00t,li2024cogact,kawaharazuka2025vlareview}. But computer-use agents faced exactly this problem only two years ago. A screen of a thousand by a thousand pixels is effectively a continuous space, and early agents could not act in it precisely on their own, relying instead on Set-of-Mark overlays~\citep{yang2023som,xie2024osworld} or separate grounding models~\citep{gou2025uground}. Yet with enough training, neither turned out to be necessary: today, a single model operates computers by simply writing click coordinates as ordinary text tokens, without any special head or tokenizer~\citep{cheng2024seeclick,qin2025uitars,bai2025qwen25vl}.

We therefore believe that embodied AI will follow the same path, and that specialized action heads will go the way of specialized grounding models~\citep{goyal2025vla0,hancock2026vlm2vla}. In fact, RT-2 and OpenVLA already show that robot actions can be expressed as tokens in the language model's own vocabulary, without any action head~\citep{brohan2023rt2,kim2024openvla}. To support this claim, we perform a preliminary investigation in Gym-Anything for Robotics~\citep{aggarwal2026gymanythingrobotics}. We train $\pi_{0.5}$~\citep{black2025pi05} and a standard vision-language model (Qwen3.5) with 9B parameters on the same 1{,}000 demonstrations for two tasks in a gift box simulation with two robot arms. \textit{We add no action head and do not even change the tokenizer.} The only change is the format in which actions are written as text, similar to computer-use agents. Yet this plain vision-language model matches $\pi_{0.5}$, succeeding in 56\% of attempts, against 59\% for $\pi_{0.5}$ across 1000 unseen instances. The model is also fast. As with $\pi_{0.5}$, the model receives a new camera image every second and either writes an action, which can span a longer duration, or chooses not to act. The 9B model takes about 400 milliseconds on average to generate an action on a single L40S GPU. Because each action can span a longer duration, decoding takes up only about 20\% of the total time and prefill about 10\%, with the remaining 70\% idle. However, this setting has only two robot arms. More complex robots, such as humanoids, have many more motors, and writing every state and action for them as plain text would make each action slower to generate. Such robots still need a more efficient format for states and actions, and possibly a better tokenizer~\citep{pertsch2025fast,kawaharazuka2025vlareview}. And even if we use additional action heads, or whatever that format turns out to be, the most important thing is, embodied AI will end up exactly where computer use did, with a single model that writes code, solves mathematics problems, operates computers, and controls robots~\citep{yang2025magma,reed2022gato,gemini2025robotics}.

\textbf{Claim 3: The path to embodied generalists follows the same playbook as computer-use agents}

What enabled the progress of computer-use agents over the last two years? We view the development of computer-use agents in three phases, each with a direct counterpart in embodied AI.

The first phase was visual grounding. Digital interfaces provide a way to generate supervision automatically: the underlying software can expose the identities and coordinates of interface elements, which can then be paired with screenshots. This approach has been used to generate grounding data at scale~\citep{gou2025uground}. Simulators offer the same opportunity for embodied AI: they can expose the identities and 3D positions of objects, which can be paired with rendered views to generate grounding data at scale~\citep{yuan2024robopoint}. Of course, this requires simulations with high enough visual fidelity, an area where coding models are clearly and continuously improving.

The second phase was building environments for reinforcement learning, following concurrent approaches used in other domains such as coding~\citep{pan2025swegym,yang2025swesmith}. Automated pipelines made it possible to generate tasks, initialize environments, and verify outcomes at scale~\citep{aggarwal2026gymanything,wang2026cuagym}. These environments then let agents learn through reinforcement learning at scale. For embodied AI, this means creating such environments in the form of simulations with realistic physics, with tasks and verifiers available through software. We further argue that embodied environments have an even stronger asymmetry between task creation and verification~\cite{aggarwal2026gymanythingrobotics}, which can be used to provide more accurate rewards or to increase task difficulty over time. We also note that reinforcement learning benefits from a strong initial policy, and training such a policy requires trajectories for pretraining or midtraining. We are therefore not advocating against teleoperation, whether in the real world or in simulation, or against egocentric videos, although this data may matter mainly in pretraining and midtraining, as in the training recipes of existing LLMs. How much of this data is needed also remains unclear, but based on preliminary results presented in this paper, we believe that perhaps much of it will not be needed, and that research on environments for reinforcement learning will remain more important.

The third phase is speed and real-time interaction, which is especially important for robotics. An agent must choose and execute actions quickly enough to keep up with changes in its environment. This requires optimizing completion time alongside task success~\citep{aggarwal2026cuaspeedrun}. Even without such optimization, the 9B model in Claim~2 took about 400 milliseconds per action and was idle for 70\% of the time. We can pursue this through rewards for completing tasks successfully with less reasoning or in less time~\citep{aggarwal2025l1,arora2025efficient}, faster inference hardware~\citep{lie2023cerebras,abts2020groq,he2025waferllm}, and harnesses that interleave quick actions with slower planning~\citep{shi2025hirobot,nvidia2025gr00t}. Such agentic harnesses organize how an agent plans, acts, and checks its progress~\citep{openai2026computeruse,koh2026macu}, and embodied tasks can reuse them once they are exposed through software.

The sequence is therefore the same: learn grounding, build environments for reinforcement learning, and optimize for real-time interaction.

\textbf{So what should be next steps for Embodied AI?}

We therefore advocate for 3 research directions we believe would become fundamental for embodied AI research: a.) improving visual grounding through data generated automatically from simulators~\citep{pfaff2026scenesmith,yuan2024robopoint}, b.) building robotic environments as simulations for robotics, exposed through software that computer-use agents can operate and be RL trained on~\citep{aggarwal2026gymanythingrobotics,wang2024robogen,hua2024gensim2}, and c.) optimizing agents for real-time interaction~\citep{aggarwal2026gymanythingrt,tang2026realtimebench} through architecture research~\citep{choudhury2024rlt,tang2026onevision,aggarwal2026vtr} and by rewarding speed alongside task success~\citep{arora2025efficient,aggarwal2025l1}. These directions call on embodied AI researchers to expose their simulators and robots through software, and on developers of computer-use agents to include such software in training.

\section{Conclusion}

In this paper, we argued that computer-use agents can already act as embodied generalists, because embodied tasks can be formulated as interaction with digital software. Using a single computer-use agent and changing only the software, we won prizes in four ECCV workshop competitions, from navigation to video segmentation. Operating that software through a GUI cost little compared to a text API, and a vision-language model with no action head matched $\pi_{0.5}$ when trained on the same robot demonstrations. We then traced how computer-use agents became general and superhuman over the last two years, through visual grounding, environments for reinforcement learning, and speed, and argued that embodied AI should follow the same playbook. We believe embodied AI is headed the same way, and that its generalist version will be the same model that already writes code, solves mathematics problems, and operates digital computers.

\section*{Acknowledgements}

We thank Daytona for their generous support in sandbox compute credits, and Google for their generous support through a Gemini API research grant. We thank the organizers of the VLNVerse, DialNav, OpenSUN3D, and MeViSv2 competitions and their ECCV 2026 workshops. All opinions expressed in this paper are the authors' own and do not necessarily reflect the views of our sponsors or the competition organizers.

\end{document}